\documentclass{article}
\usepackage{spconf,amsmath,graphicx}
\usepackage{enumitem}
\usepackage{url}
\setlist{nosep, leftmargin=14pt}
\usepackage{mwe}
\usepackage{hyperref}  

\title{Automated Prediction of Paravalvular Regurgitation before Transcatheter Aortic Valve Implantation}

\name{\begin{tabular}{c} Michele Cannito$^{1}$ \quad Riccardo Renzulli$^{1}$ \quad Adson Duarte$^{1}$ \quad Farzad Nikfam$^{1}$ \\ \quad Carlo Alberto Barbano$^{1}$ \quad Enrico Chiesa$^{1}$ \quad Francesco Bruno$^{1}$ \quad Federico Giacobbe$^{1}$ \\ \quad Wojciech Wanha$^{2}$ \quad Arturo Giordano$^{3}$ \quad Marco Grangetto$^{1}$ \quad Fabrizio D'Ascenzo$^{1}$\end{tabular}}

\address{
$^{1}$University of Turin, Italy \quad $^{2}$Medical University of Silesia, Poland \quad $^{3}$Pineta Grande Hospital, Italy
}

\begin{document}
\ninept
\maketitle

\begin{abstract}

Severe aortic stenosis is a common and life-threatening condition in elderly patients, often treated with Transcatheter Aortic Valve Implantation (TAVI). Despite procedural advances, paravalvular aortic regurgitation (PVR) remains one of the most frequent post-TAVI complications, with a proven impact on long-term prognosis.

In this work, we investigate the potential of deep learning to predict the occurrence of PVR from preoperative cardiac CT. To this end, a dataset of preoperative TAVI patients was collected, and 3D convolutional neural networks were trained on isotropic CT volumes. The results achieved suggest that volumetric deep learning can capture subtle anatomical features from pre-TAVI imaging, opening new perspectives for personalized risk assessment and procedural optimization. Source code is available at \url{https://github.com/EIDOSLAB/tavi}.
\end{abstract}



\section{Introduction}
\label{sec:intro}

Aortic stenosis (AS) is the most common valvular heart disease in the elderly, characterized by progressive narrowing and stiffening of the aortic valve leaflets due to fibrosis, calcification, and endothelial damage~\cite{ otto_calcific_2024}. 
Transcatheter Aortic Valve Implantation (TAVI) is a minimally invasive approach to treat severe symptomatic AS, especially in patients at high or intermediate surgical risk~\cite{clayton_transcatheter_2014}. While TAVI has revolutionized the treatment of patients with severe aortic stenosis, offering a less invasive alternative for those at high or prohibitive surgical risk, the procedure is still associated with several periprocedural and long-term complications ~\cite{desai_transcatheter_nodate, kapadia_long-term_2014}.

Among these, \textit{paravalvular aortic regurgitation} (PVR) remains one of the most clinically relevant, even with the latest generations of prosthetic devices~\cite{jilaihawi_paravalvular_2016}. Multiple studies have demonstrated that moderate or severe PVR significantly worsens survival, and even mild regurgitation has been associated with increased long-term mortality following TAVI~\cite{yokoyama_impact_2023}. Preoperative identification of patients at higher risk of PVR is therefore crucial for optimal valve selection and procedural planning. However, predicting post-procedural regurgitation remains challenging, as it depends on a complex interplay of anatomical and procedural factors, including annular geometry, degree and distribution of calcifications, and prosthesis–patient mismatch. Current workflows rely primarily on manual CT measurements, which provide limited insight into these multidimensional relationships and may overlook subtle 3D or textural patterns associated with regurgitation risk~\cite{salgado_preprocedural_2014, annoni_pre-tavi_2023}.
Recent advances in deep learning have enabled 3D convolutional neural networks to learn complex volumetric representations from medical images, showing promising applications in TAVI planning and anatomical analysis~\cite{saitta_ct-based_2023}. However, even for experienced clinicians, identifying subtle imaging patterns predictive of paravalvular regurgitation in preoperative CT remains challenging due to the subtle and multifactorial nature of the condition. Building on this premise, we propose a 3D deep learning framework to investigate whether relevant spatial and textural features associated with PVR can be automatically extracted from pre-TAVI CT volumes. The framework compares multiple DL architectures and examines the effect of pretraining and anatomical segmentation on model generalization.
\section{Related Works}
\label{sec:prevworks}

Most existing approaches focus on post--procedural assessment or outcome prediction rather than exploiting preoperative imaging for risk stratification. Deep learning models have been applied to automate pre--TAVI CT analysis, enabling accurate extraction of aortic annulus geometry and coronary distances, showing strong agreement with manual measurements~\cite{wang_development_2023}. Similarly, machine learning pipelines using preprocedural CT angiography have been explored for predicting all--cause mortality after TAVI, achieving higher predictive power than clinical risk scores~\cite{kwiecinski_preprocedural_2025}. Other works proposed fully automatic 3D neural networks for aortic root segmentation and valve sizing, demonstrating the feasibility of end--to--end automation in clinical workflows~\cite{wang_development_2023}. Despite these advances, few studies have leveraged deep learning on \emph{preoperative CT volumes} to predict specific post--TAVI complications such as paravalvular regurgitation (PVR). Existing AI applications primarily address anatomical quantification, post--procedural imaging, or device positioning leaving the predictive potential of pre--TAVI CT largely unexplored.

Analogous efforts in other surgical domains highlight the promise of deep learning on preoperative CT for predictive modeling. Examples include survival prediction in lung cancer from preoperative chest CT~\cite{kim_preoperative_2020}, complexity estimation in abdominal wall reconstruction~\cite{elhage_development_2021}, and lymphovascular invasion prediction from contrast--enhanced gastric CT~\cite{sun_preoperative_2025}. These works collectively demonstrate that volumetric deep learning can extract subtle preoperative imaging biomarkers associated with post--procedural risk.

Building on this evidence, our study focuses on the use of 3D convolutional neural networks trained directly on preoperative CT volumes to predict PVR risk after TAVI. To our knowledge, this represents one of the first attempts to investigate whether morphological and textural patterns embedded in pre--TAVI CT can contribute to personalized risk prediction before valve implantation.

\section{Materials and Methods}
\label{sec:methods}

\subsection{Dataset}
The study included preoperative cardiac CT scans from 249 patients undergoing Transcatheter Aortic Valve Implantation (TAVI) at A.O.U. Città della Salute e della Scienza di Torino. Each examination comprised multiple cardiac phases acquired from gated CT angiography covering the entire thoracic region, of which the first phase was selected as the most representative for analysis.
 
Images were reconstructed at a resolution of $512\times512$ pixels with an average in-plane spacing of $0.49 \pm 0.06$~mm and slice thickness of $0.625$~mm, all acquired by the same device.  
Among the available patients, 174 presented no paravalvular regurgitation and 75 presented mild or moderate-to-severe regurgitation, leading to a slightly imbalanced dataset from real world cases. Patients were divided into training and test subsets using a stratified sampling strategy applied at the patient level. 
The final split included 199 patients for training and 50 for testing, consistent with the overall class balance of the dataset.

\subsection{Preprocessing}
All CT volumes were preprocessed following a standardized pipeline designed to ensure uniformity across patients and cardiac phases.  
Each slice was resized to $256\times256$ pixels and resampled to isotropic voxel spacing of $0.625~\mathrm{mm}^3$ using 3D interpolation.  
Volumes were then padded or cropped along the axial direction to obtain a fixed grid of $256\times256\times256$ voxels. Voxel intensities were clipped to the range \([-1000, 2000]\) Hounsfield Units (HU), which effectively preserves all physiologically relevant tissues (air, soft tissue, bone, and contrast-enhanced regions) while excluding artifacts and non-informative extremes.  
Compared to the previous approach without clipping, this normalization significantly enhanced tissue contrast and local detail visibility by concentrating the gray-scale dynamic range within meaningful HU intervals. Data augmentation was applied exclusively to the training set to improve generalization and mitigate overfitting.  
Specifically, random 3D rotations between $10^\circ$ and $20^\circ$ were used, preserving anatomical coherence and simulating minor patient positioning variability.

To assess the impact of anatomical focus on model performance, an additional series of experiments was conducted using CT volumes cropped according to automatic segmentations of the heart and aorta generated with TotalSegmentator \cite{wasserthal_totalsegmentator_2023}. This approach aimed to determine whether restricting the field of view to clinically relevant regions could enhance discriminative capability by reducing background variability and emphasizing morphologically informative structures.

\subsection{Architectures and Pretraining}
We evaluated four architectures for volumetric classification of preoperative CT scans: two convolutional baselines and two pretrained models. 
The baseline networks consisted of a 3D DenseNet121 \cite{huang_densely_2018} and a 3D ResNet50 \cite{he_deep_2016}, both implemented using the MONAI framework and adapted to process single-channel volumetric inputs of size $256^3$. 
Each network produced a single logit representing the probability of paravalvular regurgitation (PVR).

In addition to models trained from scratch, two pretrained variants were explored, adopting a transfer learning strategy to improve feature generalization and convergence stability given the limited size of the TAVI dataset.
The first model is a DenseNet121 initialized from a network pretrained to perform \textit{coronary calcium scoring} on non-gated chest CT scans from an institutional dataset provided by A.O.U.\ Città della Salute e della Scienza di Torino, which includes 491 paired frontal chest radiographs and corresponding thoracic CT examinations of the same patients \cite{gallone_detection_2025}. 

The second pretrained model was obtained from the open-access \textit{COCA} dataset, comprising 213 non-gated chest CT scans \cite{li_combating_2023}.
Each scan was paired with a reference Agatston calcium score used as a continuous regression target to train a 3D DenseNet121 for predicting the logarithm of the total coronary calcium score.
Preprocessing was kept identical to that of the A.O.U.\ Molinette dataset to ensure compatibility of the learned representations.
Although COCA scans are more heterogeneous and of lower resolution (mean slice thickness $\approx$ 5 mm), they capture realistic thoracic anatomy and calcification patterns, providing transferable morphological features potentially linked to post-TAVI paravalvular regurgitation.

\subsection{Loss and Metrics}
Two loss functions were evaluated to optimize binary volumetric classification: Binary Cross-Entropy (BCE) and Focal Loss. 
The BCE loss is defined as
\begin{equation}
\mathcal{L}_{\mathrm{BCE}} = - \frac{1}{N} \sum_{i=1}^{N} [ y_i \log(\hat{y}_i) + (1 - y_i) \log(1 - \hat{y}_i) ],
\end{equation}
where $y_i$ and $\hat{y}_i$ denote the ground truth and predicted probability for sample $i$, respectively. 
The Focal Loss extends BCE by introducing a modulation factor that reduces the relative loss for well-classified examples, focusing training on harder cases:
\begin{equation}
\mathcal{L}_{\mathrm{Focal}} = - \frac{1}{N} \sum_{i=1}^{N} (1 - \hat{y}_i)^{\gamma} y_i \log(\hat{y}_i) + \hat{y}_i^{\gamma} (1 - y_i) \log(1 - \hat{y}_i),
\end{equation}
with $\gamma$ controlling the degree of focusing.

Model performance was primarily assessed using Balanced Accuracy (BA) which provides a fair estimate of performance under class imbalance by equally weighting sensitivity and specificity.

\subsection{Model Interpretability}
\begin{figure}[!t]
\centering
\includegraphics[width=0.9\linewidth]{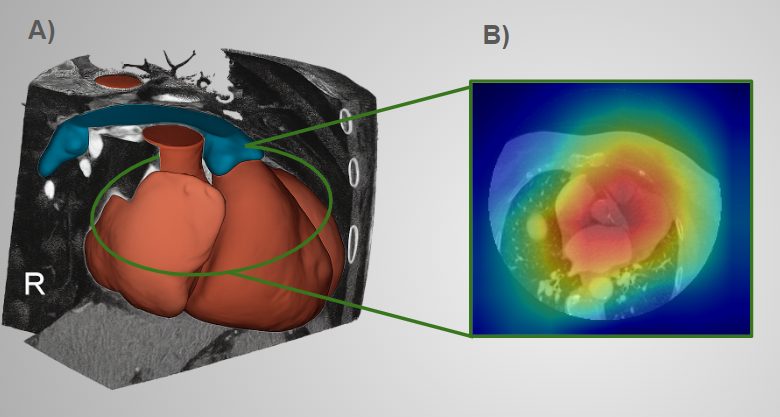}
\caption{
Overview of the proposed pipeline and model performance. 
\textbf{A)} 3D rendering of a representative preoperative CT showing the automatically segmented heart (red) and aorta (blue). 
\textbf{B)} Grad-CAM activation map. 
}
\label{fig:overview_results}
\end{figure}

To qualitatively verify whether the network focused on anatomically relevant regions during prediction, three-dimensional Gradient-weighted Class Activation Mapping (3D~Grad-CAM) was employed on a random subset of test patients. 
Activation maps were generated from the last convolutional layer of the trained models and overlaid on the original CT volumes to visualize areas contributing most to the predicted outcome. To provide a broader visual overview of the study, Figure~\ref{fig:overview_results} integrates anatomical visualization, model interpretability, and predictive performance of the best-performing network.

\section{Results}
\label{sec:results}

\subsection{Training Setup}
An initial hyperparameter search was conducted to identify the most effective training configuration for volumetric classification.
Different combinations of loss functions, optimizers, and learning rate schedules were tested, while data preprocessing and batch size were kept fixed.
After this exploration phase, a single configuration was adopted for all subsequent experiments to ensure consistency. All models were trained using Focal Loss ($\gamma = 2$ and $\alpha = 0.75$) optimized with AdamW (constant learning rate $1\times10^{-4}$, weight decay $1\times10^{-3}$). 
Training was performed with a batch size of 8 for up to 500 epochs, using automatic mixed precision (AMP) and dynamic gradient scaling on Nvidia A40 GPUs.
Each configuration was repeated three times with different random seeds, and the mean and standard deviation of the balanced accuracy across runs were reported.

\subsection{Baseline Experiments}
The initial experiments compared two convolutional backbones trained from scratch: a 3D~DenseNet121 and a 3D~ResNet50, both adapted for single-channel volumetric inputs of size $256^3$.
Under this setting, the ResNet50 achieved a mean balanced accuracy (BA) of \textbf{66.3\%}, while the DenseNet121 reached \textbf{67.4\%}.
Despite slightly higher variability across runs, the DenseNet exhibited smoother training dynamics and better overall generalization, leading us to select it as the baseline architecture for subsequent experiments.

\subsection{Pretraining Experiments}
\begin{figure}[htbp]
    \centering
    \includegraphics[width=\linewidth]{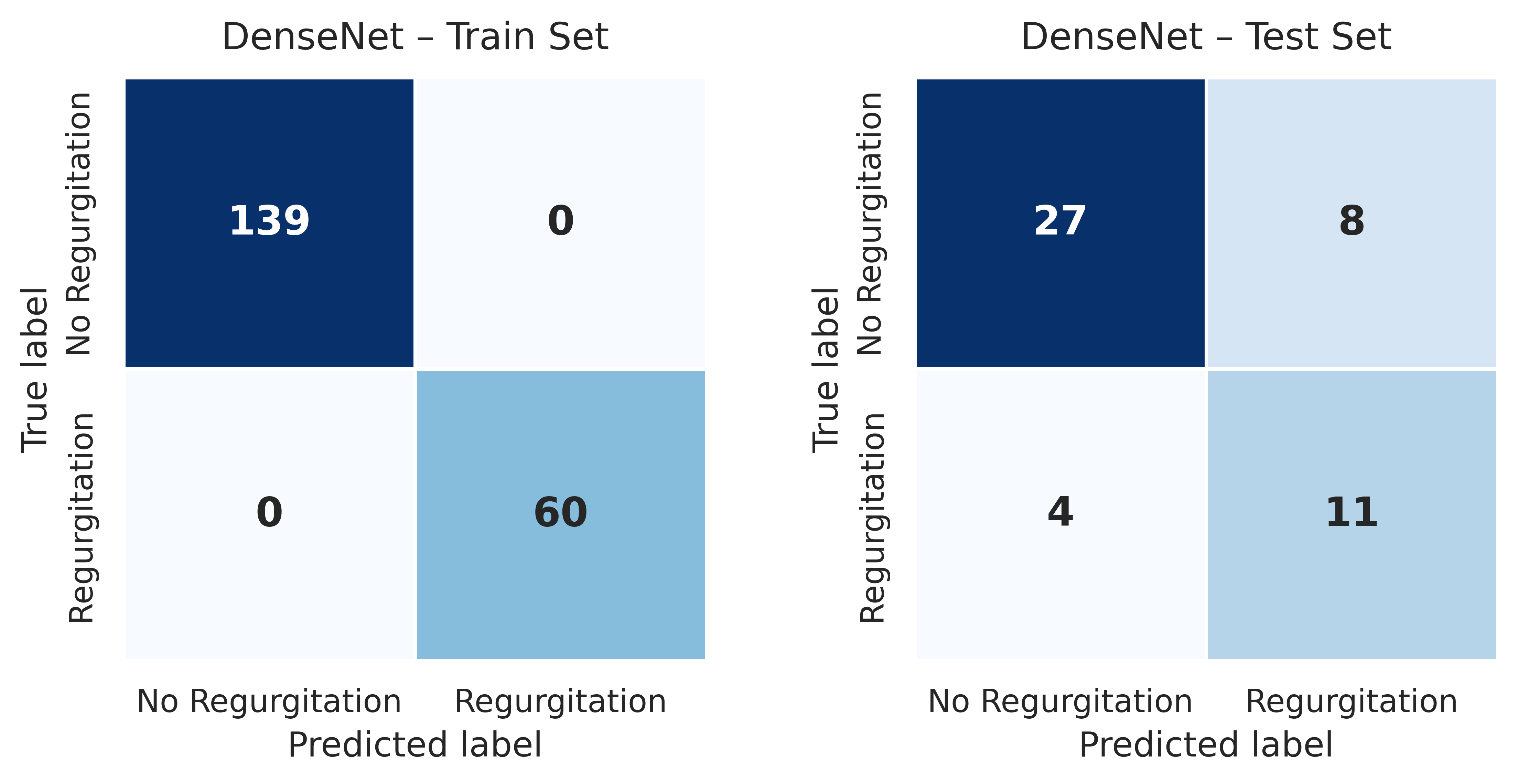}
    \caption{Confusion matrices of the DenseNet model after pretraining and fine-tuning. 
    (\textbf{Left}) Training set. (\textbf{Right}) Test set. 
    }
    \label{fig:cm_dense_pretrain}
\end{figure}

Building upon the baseline, two types of pretraining were explored. 
When pretrained on the open-access COCA dataset, the model reached a BA of \textbf{64.1\%}, showing limited transferability due to the lower quality and heterogeneity of COCA scans.
Conversely, pretraining on an internal coronary calcium scoring dataset led to a markedly higher BA of \textbf{71.8\%}, confirming the benefit of domain-specific initialization with anatomically consistent data. Figure~\ref{fig:cm_dense_pretrain} illustrates the resulting confusion matrices on the training and test sets after fine-tuning the DenseNet model. 
While the model achieves near-perfect separation on the training set, the test set reveals residual misclassifications, particularly among cases with mild paravalvular regurgitation, reflecting the intrinsic challenge of the task.

\subsection{Segmentation-based Evaluation}
To assess whether anatomical restriction could further improve discrimination, both pretrained DenseNet variants were retrained using CT volumes cropped according to automatic heart and aorta segmentations.
While mean performance slightly decreased (BA = 69.9 ± 1.0 for the internally pretrained model and 66.8 ± 1.3 for the COCA-pretrained one), the reduced variability suggests a more stable behavior across runs.
These results indicate that excluding peripheral anatomical context does not markedly degrade performance, although it may limit the availability of subtle yet informative cues.
A summary of all configurations is provided in Table~\ref{tab:results_summary}.
\begin{table}[!h]
\centering
\caption{Comparison of experimental configurations and corresponding test Balanced Accuracy (BA, mean ± SD over 3 runs). 
PT = Pretraining, FT = Fine-tuning.}
\label{tab:results_summary}
\renewcommand{\arraystretch}{1.1}
\setlength{\tabcolsep}{4pt}
\small
\begin{tabular}{p{2.2cm} p{4.3cm} c}
\hline
\textbf{Model} & \textbf{Training Strategy} & \textbf{Test BA (\%)} \\
\hline
\multicolumn{3}{c}{\textbf{Baseline (trained from scratch)}} \\
ResNet50 & Focal Loss ($\gamma=2$, $\alpha=0.75$) & 66.3 ± 1.3 \\
DenseNet121 & Focal Loss ($\gamma=2$, $\alpha=0.75$) & 67.4 ± 2.9 \\
\hline
\multicolumn{3}{c}{\textbf{PT on full CT volumes}} \\
DenseNet121 & COCA PT → TAVI FT & 64.1 ± 2.3 \\
DenseNet121 & Internal PT → TAVI FT & \textbf{71.7 ± 3.3} \\
\hline
\multicolumn{3}{c}{\textbf{Segmentation-based experiments}} \\
DenseNet121 & COCA PT + segm. input & 66.8 ± 1.3 \\
DenseNet121 & Internal PT + segm. input & 69.9 ± 1.0 \\

\hline
\end{tabular}
\end{table}

\section{Discussion and Conclusion}
\label{sec:discussion}

This study investigated deep learning approaches for predicting post-TAVI paravalvular regurgitation (PVR) using preoperative CT scans only, aiming to identify high-risk patients before intervention. 
This focus is clinically relevant, as reliable preprocedural prediction could support valve selection and procedural planning.

Among the evaluated architectures, the 3D DenseNet121 achieved slightly higher performance than the 3D ResNet50 when trained from scratch, showing smoother convergence and better generalization trends despite the limited dataset size.
The use of Focal Loss effectively mitigated class imbalance, improving sensitivity to PVR-positive cases.
Transfer learning proved to be a key determinant of performance: pretraining on an internal coronary calcium scoring task yielded the highest balanced accuracy (\textbf{71.7 ± 3.3\%}), confirming the value of domain-specific initialization on anatomically consistent thoracic CT data.
Conversely, pretraining on the COCA dataset resulted in poorer transferability, likely due to the lower image quality and greater heterogeneity of the source domain.
When restricting the input to automatically segmented heart and aorta regions, the mean performance slightly decreased (\textbf{69.9 ± 1.0\%}) but with substantially reduced variability, suggesting a more stable yet less context-aware model.
Overall, these findings reveal a trade-off between accuracy and robustness, where preserving full anatomical context enhances discriminative power at the cost of higher variance.

In conclusion, this work demonstrates the feasibility of preoperative CT-based deep learning for PVR risk prediction and underscores the importance of domain-specific pretraining and contextual information.
Future developments will aim to expand the dataset, integrate multimodal clinical features, and investigate attention-guided or hybrid segmentation strategies to further improve robustness and interpretability.

\section{Compliance with Ethical Standards}
\label{sec:ethics}
This study was performed in accordance with the Declaration of Helsinki and approved by the institutional ethics committee. All patient data were anonymized prior to processing.


\bibliographystyle{IEEEbib}
\bibliography{refs.bib}

\end{document}